\pdfoutput=1

\documentclass[11pt]{article}

\usepackage{naacl2021}
\usepackage{mathrsfs}   
\usepackage{float}
\usepackage{times}
\usepackage{latexsym}
\usepackage{amsmath}
\usepackage{graphicx}
\usepackage{booktabs}
\usepackage{multirow}
\usepackage[T1]{fontenc}

\usepackage[utf8]{inputenc}
\usepackage{fvextra}
\usepackage{microtype}

\title{Inference-Time Steering for Cross-Lingual Factual Consistency in LLMs}

\author{Alexander Manev \\
        Technical University of Munich \\
        \texttt{alexander.manev@tum.de}}

\begin{document}
\maketitle
\begin{abstract}
Although Large Language Models (LLMs) demonstrate remarkable multilingual fluency, their internal knowledge representations remain disproportionately biased toward high-resource languages. This leads to cross-lingual factual inconsistency, where they shift their empirical answer distributions based solely on the prompt language. We investigate whether these biases can be mitigated at inference time, forcing an English-prompted model to answer as if it were queried in target languages (German, Spanish, Bulgarian), and evaluate four intervention strategies: zero-shot contextual steering (persona prompting), internal representation manipulation via Contrastive Activation Addition (CAA), and lightweight weight modification via Direct Preference Optimization (DPO) trained on benchmark-derived factual data as well as conceptual generalization data. To assess alignment, we curate a multilingual factual dataset alongside a novel generalization benchmark comprising culturally rooted queries to determine whether factual interventions transfer to broader target-centric preferences. Experiments on Gemma 3 12B Instruct reveal persona prompting to be the strongest overall intervention, balancing efficacy, safety, and out-of-domain generalization. While CAA yields sharp inconsistency benchmark shifts, it is configuration-sensitive and risks knowledge degradation. DPO-based adapters offer permanent, yet narrower and less transferable gains. These findings suggest that cross-lingual inconsistency is at least partly a selection problem, and that simple contextual interventions may outperform more invasive methods for robust, transferable alignment.
\end{abstract}

\section{Introduction}

The ability of Large Language Models (LLMs) to store and recall vast amounts of factual knowledge across hundreds of languages has been a cornerstone of their global adoption. However, despite their multilingual fluency, their internal knowledge representations remain fundamentally biased toward high-resource languages, primarily English. Consequently, factual recall in modern LLMs is often highly inconsistent across linguistic contexts. When queried about a fact with multiple valid answers---such as the nationality of a public figure with dual citizenship---an LLM may exhibit a strong preference for the answer most salient in the English-speaking world. Yet, when prompted about the exact same entity in a different language, the model's empirical answer distribution frequently undergoes a drastic shift. This phenomenon of cross-lingual factual inconsistency reveals representational biases shaped by uneven training data, severely undermining the reliability, equity, and trustworthiness of LLMs for a global user base.

While extensive research has focused on mitigating these biases during pre-training or massive instruction fine-tuning, deeper structural biases inevitably persist in the final weights. This raises a critical question for model alignment and mechanistic interpretability: if multilingual representations share a common semantic space, can we actively intervene during generation to influence the model's default cultural bias? Specifically, can we successfully force an English-prompted model to answer as if it were being queried in a target language? 

To address this, we investigate the mechanisms of cross-lingual factual consistency at inference time. Rather than relying on computationally prohibitive full-parameter fine-tuning, we design a rigorous comparative study across progressively distinct depths of intervention---ranging from zero-shot contextual steering to internal activation manipulation and lightweight parameter updates.
Crucially, any targeted intervention risks degrading the model's behavior elsewhere. Hence, we evaluate these strategies not merely on shifting specific probabilities, but also for safety on ordinary facts and for transfer to a broader out-of-benchmark setting.

In summary, our primary contributions are:
\begin{itemize}
    \item \textbf{A multilingual factual dataset:} a curated high-quality dataset of factual triples across English, German, Spanish, Bulgarian, Swedish, and Korean, designed to capture cross-lingual distributional shifts under repeated non-zero-temperature sampling.
    \item \textbf{A distributional evaluation framework:} an extraction-based analysis showing that multilingual factual behavior remains substantially inconsistent across languages even when the underlying model is held fixed.
    \item \textbf{A novel generalization benchmark:} a culturally rooted dataset (50 queries per language) to test whether interventions transfer beyond benchmark-specific factual inconsistencies.
    \item \textbf{A comparative intervention study:} an evaluation of prompting, CAA, benchmark- and generalization-derived DPO showing that prompting is the strongest overall, while more invasive methods provide useful but more limited gains in robustness and transfer.
\end{itemize}

\section{Related work}


The disparity in factual knowledge representation across languages remains a critical vulnerability in modern LLMs, which disproportionately favor English-centric knowledge \citep{shafayat2024multifact, shcharbakova2025cross}.
This asymmetry is frequently analyzed through the \textit{Semantic Hub Hypothesis} \citep{wu2025semantichubhypothesis}, which posits that models map multilingual inputs into a shared, language-agnostic conceptual space in their intermediate layers before verbalizing them in the target language. Because this shared space is heavily anchored by English pre-training distributions, failures in the final ``language transition'' step cause models to act as distinct cultural entities depending solely on the language of the prompt \citep{wang2025lostmultilinguality}.
To measure whether models genuinely understand target cultures or merely translate surface-level text, the community has developed specialized evaluation frameworks like FORK \citep{palta2023fork}, NormAd \citep{rao2025normad}, CulturalBench \citep{chiu2025cultural}, and XLQA \citep{roh2025xlqa}. 

While these state-of-the-art datasets excel at the static evaluation of foundational models, they are not designed to measure the depth of post-hoc intervention alignment. Our work introduces a novel generalization benchmark utilizing highly localized, second-person scenarios formulated entirely in English. This design isolates the intervention's effect, allowing us to systematically test whether a steering method induces a genuine conceptual ``worldview'' shift within the model's semantic hub.

Recent efforts to mitigate these representational bottlenecks predominantly rely on resource-intensive training interventions, such as batch-wise fine-tuning for structural consistency \citep{agarwal2025aligning}, representation forcing \citep{bu2025alignx}, multilingual alignment scaling \citep{shen2025unaligned}, and truthfulness transfer from high-resource languages \citep{liu2025selected}. The underlying assumption of these works is that the model's semantic space must be permanently overwritten or aligned via parameter updates. In contrast, our work challenges this assumption: we hypothesize that the target cultural knowledge already exists within the model's weights and investigate whether an English-prompted model can be actively steered at \textit{inference time} to dynamically adopt distinct empirical answer distributions.

To control model outputs without the prohibitive computational costs of full-parameter fine-tuning, researchers increasingly rely on activation engineering. Building upon foundational representation engineering techniques \citep{zou2025representation, turner2024steering}, \textbf{Contrastive Activation Addition (CAA)} \citep{panickssery2024steeringllama} allows for direct intervention in a model's internal state via \textit{steering vectors}. 
While CAA has been successfully used to reduce sycophancy or enforce general truthfulness, we are, to the best of our knowledge, the first to apply it to cross-lingual preference realignment. Crucially, instead of using externally curated behavioral examples, we derive our steering vectors directly from the model's own baseline distributions, attempting to force the English residual stream to mimic the target language's activation state.

We juxtapose this mechanistic intervention with \textbf{persona prompting}, a zero-shot contextual baseline. Recent work demonstrates explicit sociodemographic persona prompting can significantly shift a model's value alignment and cultural representation \citep{wang2025multilingual}. However, studies also warn that while effectively steering subjective generation, it can be unpredictable and occasionally detrimental to objective factual recall \citep{zheng2024ahelpfulassistant, lutz2025prompt}. Utilizing this technique tests the limits of the model's in-context retrieval---whether it can reliably adopt a target demographic's perspective without altering internal activations.

While our primary focus is inference-time interventions, we rigorously benchmark these against permanent parameter updates via a modern post-training alignment technique: \textbf{Direct Preference Optimization (DPO)} \citep{rafailov2024directpreferenceoptimization}. Although recent works adapt preference optimization for inference-time decoding \citep{chen2025diffpo}, we use standard DPO and re-purpose it: rather than correcting factually ``wrong'' outputs or enforcing universal safety stances, we train lightweight LoRA adapters on contrastive pairs where neither option is strictly incorrect but merely culturally preferred. This reframes cross-lingual consistency as a targeted preference learning problem, allowing us to contrast representation collapse risks of DPO weight modification \citep{pal2024smaug} against the flexibility of inference-time approaches.

\section{Data}


We evaluate our interventions on two complementary datasets. The first is a \textbf{multilingual factual benchmark} designed to expose cross-lingual distributional shifts in open-ended factual recall. Building upon \texttt{KLAR} \citep{wang2025lostmultilinguality}, we curated a carefully fact-checked benchmark of atomic triples (subject, relation, object), thereby reducing the influence of prompt context on model outputs. This dataset spans six typologically diverse languages---English, German, Spanish, Bulgarian, Korean, and Swedish---and covers seven relations specifically chosen to elicit both standard recall and culturally sensitive variation. The benchmark is designed for repeated-sampling evaluation of answer distributions rather than single-shot exact-match accuracy; full curation details, relation choices, and format examples are provided in Appendix~\ref{sec:app_dataset_details}.

Our second resource is a novel \textbf{generalization dataset} targeting broader cultural transfer beyond benchmark-specific facts. It contains 50 culturally rooted multiple-choice scenarios for each target language (German, Spanish, Bulgarian), all formulated in English and framed in the second person. This design probes whether an intervention merely shifts benchmark-local factual preferences or successfully induces a broader target-centric perspective under English prompting. Construction details and example items are given in Appendix~\ref{sec:app_generalization_construction}.

\section{Baseline}
Before intervening, we first measured the extent of cross-lingual factual inconsistency across four state-of-the-art LLMs: Gemma 3 12B Instruct \citep{gemmateam2025gemma3}, gpt-oss-20b \citep{openai2025gptoss120bgptoss20bmodel}, Qwen 3 30B-A3B \citep{yang2025qwen3}, and Llama 4 Maverick Instruct \citep{meta2025llama4}. To bypass the biases of multiple-choice recognition, we used open-ended generation under repeated sampling. For each subject-relation pair, we generated 10 samples at each of $T\in\{0.8,1.2\}$ with randomized prompt templates, treating answer frequency as a proxy for probability mass. Outputs were post-processed with an extraction layer that canonicalized answers into English, allowing us to compute empirical answer distributions and quantify cross-lingual inconsistency using \textbf{pairwise Jensen--Shannon distance (JSD)}. A separate judgment layer was implemented for correctness-oriented follow-up analysis (full generation and evaluation details in Appendix~\ref{sec:app_baseline_details}).

The baseline confirmed that multilingual factual behavior is systematically unstable. Mean pairwise JSD ranged from 0.1258 (Spanish--Swedish) to 0.2818 (Bulgarian--Korean), with the strongest divergence appearing in culturally fluid categories like \textit{occupation} (0.4051) and \textit{city of origin} (0.3972), and the weakest in more constrained facts like \textit{capital} (0.1253). These findings motivated our intervention focus on German, Spanish, and Bulgarian, which preserve representative typological diversity while spanning a useful range of cross-lingual divergence (detailed baseline findings and target-language selection rationale in Appendix~\ref{sec:app_baseline_findings}).

\section{Interventions}\label{sec:interventions}
Having established this baseline, we designed a comparative experimental setup to test how deeply we must intervene to align English generations with target language preferences. All experiments were conducted on Gemma 3 12B Instruct; the model-selection rationale is provided in Appendix~\ref{sec:app_model_selection}.

\subsection{Prompting}
As the shallowest level of intervention, we first investigated whether the model already possesses the target cultural knowledge, but simply defaults to an Anglo-centric worldview due to the English prompt.
To test this, we employed zero-shot \textbf{persona prompting}. For each query, we augmented the prompt with a natural language instruction forcing the model to adopt the perspective of a demographic corresponding to the target language.
This tested the limits of the model's in-context learning and contextual retrieval capabilities before advancing to deeper mechanistic interventions. To ensure a thorough comparison, these generations were evaluated using the exact same repeated sampling methodology as the baseline. The persona prompt template utilized is provided in Appendix~\ref{sec:app_persona_prompts}.

\subsection{Steering}\label{sec:caa}
Building on our baseline findings, we next employed CAA, a method for steering model behavior by manipulating its activations \citep{panickssery2024steeringllama}. Unlike prompting, which operates purely through instruction context, CAA targets the model's hidden-state geometry directly by adding a directional bias vector to its residual stream during inference (visually outlined in Figure~\ref{fig:app_steering}).

The core of our steering methodology is the creation of a \textit{steering vector}---a directional representation of a specific behavior within the model's activation space derived from \textit{contrast pairs}. Rather than hand-curating isolated inconsistencies, we built these pairs automatically from the baseline extraction outputs. Specifically, for subjects where the answer distributions differed, we paired an English-preferred candidate (the \textit{negative} completion to be suppressed) with a target-preferred candidate (the \textit{positive} completion toward which the model should move). This operationalizes steering as \textit{preference realignment} rather than error correction, as the English-side candidate need not be factually wrong (see Appendix~\ref{sec:app_contrast_pairs} for examples).
To identify the most effective representational shift, we experimented with three contrast pair strategies:
\begin{itemize}
    \item \texttt{v0}: strict disjoint contrasting (pairs only English-only and target-only candidates).
    \item \texttt{v1}: full cross-product contrasting, excluding identical pairs.
    \item \texttt{v2}: full cross-product contrasting, including identical pairs (retains the complete English $\times$ target candidate space).
\end{itemize}

For each resulting pair, we generated an English multiple-choice prompt with randomized candidate ordering to avoid positional bias. 
A formal mathematical definition of the steering vector computation and the exact mechanics of its inference-time application are detailed in Appendix~\ref{sec:app_caa_computation}.

\subsection{Adapting}
While persona prompting and activation steering intervene only at inference time, our third strategy modifies the model parameters themselves. The motivation is straightforward: if cross-lingual factual preferences are systematic rather than incidental, then they may be learnable through preference optimization. Instead of nudging the model at generation time, we hence asked whether a small language-specific adapter can be trained to prefer the target-language answer distribution even when the prompt remains in English.

To this end, we trained parameter-efficient LoRA adapters (targeting the standard attention and feed-forward projection modules) on top of Gemma 3 using the TRL DPO trainer \citep{rafailov2024directpreferenceoptimization}. Conceptually, the objective is to increase the relative likelihood of the target-preferred continuation (\textit{chosen}) over the English-preferred one (\textit{rejected}) for the same prompt, encouraging the English-prompted model to internalize the target-language preference pattern rather than merely simulating it in context.
We trained two distinct adapter versions using identical hyperparameters (detailed in Appendix~\ref{sec:app_dpo_training}), varying only the training data. This ensures that any difference between the two can be attributed primarily to the source of preference data rather than the optimization method:

\paragraph{Benchmark-derived DPO:}
Here, the training data was derived directly from the steering pipeline. We reused the English-vs.-target contrast pairs across the three strategies described above (\texttt{v0}, \texttt{v1}, \texttt{v2}). In this way, DPO and CAA rely on the exact same underlying preference signal, differing only in how that signal is used. While steering asks whether the desired perspective can be induced by manipulating hidden activations at inference time, DPO asks whether the same preference shift can be learned as a stable, localized parameter update.

\paragraph{Generalization-derived DPO:}
Instead of deriving preferences directly from the factual benchmark itself, we asked whether the same cross-lingual shift can be induced indirectly through the independently constructed generalization dataset (\S\ref{sec:app_generalization_construction}).
Here, the culturally target-centric option served as the \textit{chosen} continuation, and the English-centric option as the \textit{rejected} response. The resulting adapter is therefore trained to internalize a broader, more abstract cultural signal rather than one tied narrowly to the benchmark subjects.

This approach goes one step further than the others, serving as both an intervention method and a transfer test. It asks whether target-language preference patterns learned in a broader synthetic setting can generalize back to factual question answering and induce the same alignment. If successful, it would suggest that cross-lingual factual alignment can be driven by a general cultural preference signal, making this the most indirect, and arguably most ambitious, intervention in our sequence.

\section{Experiments}\label{sec:experiments}
All interventions were evaluated under the same repeated-sampling generation setup, utilizing the distributional and statistical metrics established in Appendix~\ref{sec:evaluation_pipeline} (JSD to target, TPM, paired Wilcoxon signed-rank tests, CLES). Reported benchmark results are based on extraction-derived normalized answers.
For steering and both adapter-based methods, experiments were organized around the three contrast variants (\texttt{v0}, \texttt{v1}, \texttt{v2}); all individual intervention results are consolidated in Table~\ref{tab:master_results}.

\subsection{Prompting}
Prompting introduces neither trainable parameters nor activation-level modifications, yet proved highly effective. Simple contextual cuing consistently shifted English answer distributions toward the desired target, improving both JSD and TPM relative to the baseline in 8 out of 9 evaluated cases.

The gains were particularly robust for Bulgarian and German across all contrast versions.
Under \texttt{v1}, prompting significantly improved German alignment ($p<0.05$, CLES $= 0.705$), indicating that for 70.5\% of randomly paired subjects, the prompted model was strictly closer to the target distribution than the baseline. Bulgarian saw even stronger statistical consistency, achieving highly significant JSD reductions in both \texttt{v1} ($p<0.01$, CLES $= 0.672$) and \texttt{v2} ($p<0.01$, CLES $= 0.690$). 
These results demonstrate that a substantial portion of the desired cross-lingual preference shift can already be induced \textit{without} altering model weights or internal activations, making prompting a surprisingly competitive intervention given its simplicity.

\subsection{Steering}
While prompting tests contextual retrieval, steering tests mechanistic intervention in the model's forward pass. 
The effectiveness of CAA is highly dependent on tuning two interacting hyperparameters: the layer of intervention and the scalar multiplier controlling the vector's strength.
A systematic sweep across all 48 layers of Gemma and 15 multiplier values for all contrast versions (Appendix~\ref{sec:app_steering_heatmaps}) revealed that optimal configurations are highly target- and version-specific. German and Spanish favored late layers and stronger multiplier scaling, while Bulgarian preferred milder interventions, occasionally at earlier layers.

Evaluated on held-out test data using the optimally selected configurations, steering generalized effectively, improving JSD in 8 out of 9 cells. The most profound mechanistic shifts occurred within \texttt{v1}. For Bulgarian, steering substantially improved distributional alignment ($p<0.01$) with a massive effect size (CLES $= 0.741$). German similarly demonstrated a robust, statistically significant causal shift ($p<0.01$, CLES $= 0.705$).
However, the precision required for steering also exposes its fragility compared to prompting. When the preference space becomes sufficiently broad (e.g., under \texttt{v2}), fixed vectors struggle to cleanly rotate the distribution: Bulgarian's TPM regressed slightly, while Spanish worsened in JSD (0.423 to 0.433). Furthermore, in the sparser strict-disjoint setting, only Bulgarian achieved a statistically significant shift ($p<0.05$), while German and Spanish did not despite nominal gains.
Ultimately, these results confirm that steering allows cross-lingual preferences to be \textit{causally manipulated}, but its gains remain visibly more configuration-dependent than simpler contextual interventions.

\subsection{Adapting}
To maintain a controlled comparison, we fixed the TRL training across all target languages and evaluated the adapters separately under \texttt{v0}, \texttt{v1}, \texttt{v2}.

The \textbf{benchmark-derived adapter} produced clear and consistent gains, improving JSD in all 9 cases and TPM in 8 out of 9. However, statistical significance revealed a dependency on the preference structure. The most consistent causal shifts occurred exclusively within the \texttt{v1} settings, where Bulgarian and German demonstrated highly significant alignment ($p<0.01$, CLES $> 0.70$).
Conversely, when the candidate space was either too sparse (\texttt{v0}) or broadened to include noisy identical pairs (\texttt{v2}), the shifts failed to reach strict statistical significance ($p>0.05$)---despite producing massive absolute shifts, e.g., Spanish \texttt{v2} driving JSD down to 0.348 and pushing TPM to 0.837. Hence, DPO requires a rich yet strictly contrastive candidate space to learn a stable cross-lingual alignment.

The \textbf{generalization-derived adapter} successfully moved the benchmark in the correct direction, improving JSD to target in all 9 cells and increasing TPM in 8 of 9. This confirms that abstract preference signals encoded in the generalization dataset apply to factual alignment. Mirroring the behavior of the benchmark-derived adapter, the most robust transfer occurred within the strictly contrastive \texttt{v1} candidate space. Here, the generalization-derived adapter significantly reduced JSD for Bulgarian ($p< 0.01$, CLES $= 0.707$) and German ($p< 0.05$, CLES $= 0.659$). However, just like the benchmark variant, it failed to reach strict statistical significance in the noisier \texttt{v2} or sparser \texttt{v0} settings across any language ($p> 0.05$).
Furthermore, while it successfully transfers part of the preference signal, its gains were consistently weaker than those trained directly on factual contrasts. In a direct head-to-head comparison, the benchmark-derived adapter achieved superior JSD in 6 of 9 cells and superior TPM in 7 of 9. Thus, while generalized cultural preference data can successfully induce cross-lingual factual alignment, it lacks the specificity and force of benchmark-specific supervision.

Taken together, all four interventions successfully move the model toward the target language under controlled benchmark conditions. The remaining question is thus not whether cross-lingual preferences are steerable, but at what cost. Specifically, we must determine whether these gains cause collateral damage by degrading the model's behavior on non-problematic facts (\textit{safety}), and whether the induced preferences genuinely transfer to broader cultural scenarios (\textit{conceptual generalization}).

\begin{table*}[t]
\centering
\small
\begin{tabular}{ll cc|cc|cc|cc|cc}
\toprule
& & \multicolumn{2}{c}{\textbf{Baseline}} & \multicolumn{2}{c}{\textbf{Prompted}} & \multicolumn{2}{c}{\textbf{Steered}} & \multicolumn{2}{c}{\textbf{Adapted (Ben.)}} & \multicolumn{2}{c}{\textbf{Adapted (Gen.)}} \\
\cmidrule(lr){3-4} \cmidrule(lr){5-6} \cmidrule(lr){7-8} \cmidrule(lr){9-10} \cmidrule(lr){11-12}
Target & Var. & JSD$\downarrow$ & TPM$\uparrow$ & JSD$\downarrow$ & TPM$\uparrow$ & JSD$\downarrow$ & TPM$\uparrow$ & JSD$\downarrow$ & TPM$\uparrow$ & JSD$\downarrow$ & TPM$\uparrow$ \\
\midrule
bg & \texttt{v0} & 0.853 & 0.214 & 0.790\phantom{**} & 0.295 & \textbf{0.754}*~ & \textbf{0.364} & 0.830\phantom{**} & 0.255 & 0.839\phantom{**} & 0.250 \\
bg & \texttt{v1} & 0.583 & 0.662 & \textbf{0.490}** & 0.729 & 0.504** & \textbf{0.755} & 0.501** & 0.745 & 0.522** & 0.726 \\
bg & \texttt{v2} & 0.484 & 0.779 & \textbf{0.394}** & \textbf{0.848} & 0.473\phantom{**} & 0.750 & 0.479\phantom{**} & 0.750 & 0.458\phantom{**} & 0.781 \\
\midrule
de & \texttt{v0} & 0.784 & 0.358 & 0.683\phantom{**} & 0.450 & 0.636\phantom{**} & \textbf{0.467} & \textbf{0.602}\phantom{**} & \textbf{0.467} & 0.644\phantom{**} & 0.458 \\
de & \texttt{v1} & 0.524 & 0.707 & \textbf{0.396}*~ & \textbf{0.789} & 0.436** & 0.748 & 0.458** & 0.743 & 0.439*~ & 0.739 \\
de & \texttt{v2} & 0.479 & 0.775 & 0.438\phantom{**} & \textbf{0.845} & 0.421\phantom{**} & 0.789 & \textbf{0.412}\phantom{**} & 0.807 & 0.443\phantom{**} & 0.782 \\
\midrule
es & \texttt{v0} & 0.778 & 0.336 & \textbf{0.684}\phantom{**} & 0.400 & 0.686\phantom{**} & 0.386 & 0.691\phantom{**} & \textbf{0.407} & 0.747\phantom{**} & 0.350 \\
es & \texttt{v1} & 0.466 & 0.711 & \textbf{0.411}\phantom{**} & \textbf{0.752} & 0.436\phantom{**} & 0.735 & 0.462\phantom{**} & 0.715 & 0.428\phantom{**} & 0.720 \\
es & \texttt{v2} & 0.423 & 0.778 & 0.468\phantom{**} & 0.737 & 0.433\phantom{**} & 0.796 & \textbf{0.348}\phantom{**} & \textbf{0.837} & 0.413\phantom{**} & 0.776 \\
\bottomrule
\end{tabular}
\caption{Consolidated intervention results on the extracted benchmark test split. Bold values indicate the absolute best-performing intervention for that specific row and metric. Significance stars (* $p < 0.05$, ** $p < 0.01$) indicate a statistically significant shift against the unmodified baseline, meaning an intervention can be significant without being the absolute best, or be the best without reaching significance (e.g., in sparse \texttt{v0} settings).}
\label{tab:master_results}
\end{table*}

\section{Robustness and transfer}\label{sec:safety_generalization}
\subsection{Factual preservation}\label{sec:robustness}
The benchmark experiments deliberately targeted problematic subjects where the model's English and target-language answers significantly diverged. However, an intervention is not truly desirable if it improves those difficult cases at the cost of degrading the model's behavior on ordinary factual subjects that did not require correction in the first place. To evaluate \textit{safety}, for each intervention we generated outputs on the held-out full dataset (excluding the entire experiment subject union) and compared the resulting answer distributions against the corresponding baseline English generations.

We report signed metrics (\textbf{$\Delta$TPM} and \textbf{$\Delta$JSD}) to capture directional shifts, alongside \textbf{mean absolute perturbations} ($|\Delta\mathrm{TPM}|$ and $|\Delta\mathrm{JSD}|$) to quantify how strongly each intervention alters the model's normal factual behavior regardless of direction. A safe intervention should ideally leave the model nearly unchanged, yielding absolute perturbations near 0 and statistically insignificant shifts.

The safety results (Table~\ref{tab:safety_full_dataset}) reveal a different picture from the inconsistency-focused benchmark. The \textit{DPO adapters} proved to be the safest interventions: both introduce only minimal perturbation on ordinary factual subjects, as reflected in their small mean absolute shifts.
The generalization-derived adapter was the most conservative overall, confirming that training-based preference optimization doesn't catastrophically overwrite broader factual behavior.
Similarly, \textit{prompting} perturbs the model only mildly (e.g., mean $|\Delta\mathrm{JSD}|$ of 0.0034 with CLES [JSD-T] $\approx0.503$), while remaining the only intervention with both positive mean $\Delta$TPM and negative mean $\Delta$JSD. This makes it the strongest overall trade-off between benchmark effectiveness and factual safety.
Conversely, \textit{steering} produced by far the largest perturbation, oftentimes harmful. For instance, under the \texttt{v0} setting ($T=1.2$), the Bulgarian vector worsened 182 ordinary factual subjects while improving only 36, resulting in a severe distributional regression (CLES [TPM] $=0.405$). Spanish suffered a similarly destructive shift (234 worsened vs.\ 58 improved, CLES [TPM] $= 0.395$). This highlights a central limitation of activation-level control: while it can generate sharp local improvements on targeted inconsistencies, those gains do not necessarily translate into a stable intervention for broader factual use.

Overall, the safety evaluation complements the benchmark results in an essential way. While the benchmark tells us which methods can repair targeted cross-lingual inconsistencies, the held-out full dataset tells us whether they do so without damaging the model elsewhere.

\subsection{Conceptual generalization}\label{sec:generalization}
Whereas the held-out full dataset served as a \textit{safety} evaluation, our generalization dataset addresses a different question: do these interventions transfer beyond the factual benchmark to a broader set of target-vs.-English-centric cultural choices? This serves as our primary out-of-benchmark test of whether the induced preference shift captures something more general than the factual inconsistencies from which the benchmark was constructed.

To provide a comprehensive evaluation, we conducted two distinct generation settings: \textbf{stochastic} repeated sampling to compute the expected target probability mass (10 runs per subject at temperatures $T\in\{0.8,1.2\}$), and \textbf{deterministic} greedy decoding with sampling disabled ($T=0$) to isolate the model's absolute top-1 preference shift.
We compare the baseline English model with prompting, steering, and benchmark-derived adapter. For CAA and DPO, the factual benchmark configurations were carried over to test for genuine transfer. The generalization-derived adapter was omitted, as it was trained directly on this dataset.

The results (Table~\ref{tab:generalization_results}) reveal a striking mechanistic dichotomy between our interventions.
Under deterministic generation, the baseline model exhibits a strong English-centric bias, selecting the target-centric option only 32.7\% of the time on average. \textit{Prompting} changes this completely, raising the target-centric selection rate to 86.0\% across all three languages. The stochastic multirun mirrors this pattern, with prompting consistently shifting TPM from roughly one third to $\approx 86\%$. This suggests that contextual cuing acts as a broad, global persona shift, altering not just factual recall but the model's entire subjective and cultural disposition.

By contrast, \textit{steering} and the \textit{adapter} exhibit bounded, task-specific behavior. Even with the strongest benchmark configurations, deterministically both reach average target-centric choice rates of only 33.3\% and 32.7\%, respectively---virtually identical to the baseline. The stochastic evaluations confirmed this was not an artifact of greedy decoding; even at higher temperatures, both methods fail to induce a comparably broad preference shift.

This asymmetry is highly informative: prompting accesses a relatively general target-language perspective already present in the model, whereas steering and benchmark-derived DPO predominantly exploit the factual preference structure on which they were tuned.
Together with the safety evaluations, this sharpens the overall picture considerably. Prompting is not only the simplest intervention, but also the one that succeeds simultaneously on all three axes considered in this work: improving problematic benchmark cases, remaining acceptably safe on ordinary factual subjects, and transferring strongly to broader out-of-benchmark preferences. 
The more invasive methods, while causally effective, remain largely confined to the benchmark regime in which they were engineered.

\section{Discussion}\label{sec:discussion}
The results reveal a clear and, in some respects, surprising pattern. While all interventions successfully moved the model in the desired direction on the inconsistency benchmark, they differed in robustness, safety, and generalization. The strongest overall was not the most mechanistically sophisticated one, but the simplest: persona prompting.
This suggests that a substantial portion of the target-language perspective is already latent in the model's weights and can be retrieved through context alone. Consequently, the main challenge is not necessarily the \textit{absence} of a relevant cultural signal, but the model's default tendency to privilege an English-centric anchor when queried in English. In that sense, our findings provide evidence that cross-lingual inconsistency is at least partly a \textit{selection} problem rather than purely a \textit{knowledge} deficit.

By contrast, the more invasive interventions yielded a fragmented picture. Steering provided the clearest evidence that cross-lingual factual preferences are causally accessible within hidden representations. However, it was also the least robust method---requiring extensive tuning and proving highly configuration-sensitive. When applied too aggressively at an inappropriate layer, it can cause collateral damage, making the model prone to hallucinations or otherwise unstable generations. DPO adapters succeeded in packaging the preference shift into a stable parameter update, but this learned signal is relatively narrow and does not automatically scale into a broad target-language worldview.

A further nuance concerns the contrast-construction variants. For gradient- and activation-based methods (DPO and CAA), the quality of the contrastive supervision signal is paramount.
The strictly disjoint \texttt{v0} setting offered the cleanest conceptual opposition but was mathematically too sparse to yield stable updates. Conversely, the permissive \texttt{v2} provided the most data but introduced critical structural ambiguity: because it retained identical pairs, the exact same object could appear as both the ``chosen'' and ``rejected'' continuation, feeding contradictory signals to the optimizer.
The \texttt{v1} setting emerged as the optimal middle ground---balancing data volume with a strictly contrastive signal---which explains why it drove the most statistically robust causal shifts across our experiments.

Moreover, this clarifies the broader pattern. Neither steering nor the benchmark-derived adapter was trained on a rich representation of target-language social norms, habits, or cultural expectations; both were derived from a relatively small set of factual preference contrasts.
In the original work, CAA is used to control high-level model behaviors during inference (e.g., hallucination, sycophancy, refusal \citep{panickssery2024steeringllama}). Likewise, DPO was introduced as a preference-optimization method for aligning LMs to human judgments in settings such as sentiment control, summarization, and dialogue \citep{rafailov2024directpreferenceoptimization}. Pushing these methods into the nonstandard regime of cross-lingual preference realignment under English prompting reveals a clear boundary: a localized factual update does not automatically translate into stable, system-wide alignment.

Overall, our work points to a useful conceptual distinction for future alignment efforts. There is a major difference between shifting a model's answer distribution on a narrow set of known problematic facts and inducing a broader, culturally consistent perspective that is safe and generalizable.
The takeaway of this study is thus twofold: cross-lingual factual inconsistency is steerable at inference time, yet pushing that realignment beyond isolated facts towards a genuinely robust and transferable target-language perspective remains a profound challenge for mechanistic interventions---one that simple contextual cuing currently navigates effectively.

\section{Conclusion and Future work}
This project investigated whether cross-lingual factual inconsistencies can be reduced by intervening on English-prompted model behavior. 
Across increasingly invasive interventions, the results show that such inconsistencies are indeed steerable, but the manner of steering matters greatly.
Multilingual models appear to contain a substantial amount of the relevant target-language signal already; thus, retrieving it via simple contextual prompting proved effective.
More mechanistic approaches such as CAA and DPO do not automatically yield better practical outcomes---while they can successfully alter local benchmark behavior, they may not provide a broad, safe, and transferable alignment effect.

A natural direction for future work is hence to move beyond small factual contrast sets alone. To induce a robust target-language perspective rather than merely correct isolated inconsistencies, the supervision signal will likely need to combine (factual) preferences with richer behavioral, cultural, and normative information.
A particularly promising avenue is our generalization dataset itself, which already proved useful as a transfer probe. Future work could expand it substantially in scale and scope---by covering a broader range of everyday practices, social conventions, and culturally grounded expectations.
A richer and more diverse dataset would make it possible to test whether interventions can genuinely achieve such an alignment.

Finally, while our experiments focused on a dense 12B-parameter model across three target languages, our factual benchmark also includes fully verified annotations for Swedish and Korean. Extending these interventions to typologically distant, non-Latin-script languages such as Korean or testing them on structurally distinct architectures like Mixture-of-Experts (MoE) models represents a vital next step for validating the universality and limits of cross-lingual inference-time steering.

\clearpage

\section{Limitations}\label{sec:limitations}
This work has several limitations:
\begin{itemize}
    \item The intervention study was conducted on a single base model---\textbf{Gemma 3 12B Instruct}. Although this choice was justified by both multilingual quality and computational feasibility, the results should not be interpreted as model-independent. Different architectures or instruction-tuning regimes may yield different behavior for prompting, CAA, and DPO.
    \item The intervention setting was restricted to three target languages: \textbf{German}, \textbf{Spanish}, and \textbf{Bulgarian}. This subset was chosen to cover distinct typological profiles in a controlled way, but it does not span the full range of linguistic and cultural variation observed in the initial multilingual evaluation.
    Other languages may introduce qualitatively different challenges.
    \item The \textbf{CAA} and \textbf{benchmark-based DPO interventions} were derived from a relatively small set of factual preference contrasts. This signal was sufficient to induce benchmark-local shifts, but it is likely too narrow to fully represent a broader target-language perspective. The weaker generalization of these methods should therefore be interpreted in light of the limited supervision they receive.
    \item The evaluation pipeline itself relies partly on LLM-based post-processing. Both \textbf{extraction} and \textbf{judgment} were designed to combine deterministic rules with GPT-5.1-based interpretation, since exact string matching was insufficient for the scale and semantic variability of the task. While this made the evaluation substantially more robust than purely rule-based post-processing, it also introduces a source of noise: some answers may have been normalized incorrectly---especially in ambiguous cases---leading to minor inaccuracies in the reported metrics. As full manual verification was infeasible at this scale, such errors cannot be ruled out entirely.
    \item The extensive \textbf{layer-} and \textbf{multiplier-sweeps} required to find the optimal steering configurations relied purely on faster, rule-based extraction. Consequently, the selected CAA configurations might slightly deviate from the absolute theoretical optimum.
    \item The primary evaluation in this project is \textbf{distributional} rather than \textbf{correctness-centered}. While a separate judgment pipeline for factual correctness and instruction compliance was implemented, it was not the main evaluation lens here. The present work is therefore strongest as an analysis of cross-lingual answer consistency and intervention behavior, rather than as a complete study of multilingual factual correctness.
    \item The \textbf{generalization dataset} used in the later experiments provides a useful probe of target-vs.-English-centric preferences, but it cannot be treated as a complete proxy for a target-language worldview. Strong or weak performance on this dataset should therefore be interpreted as evidence about transfer of preference signals, not as a full account of cultural alignment.
\end{itemize}

\section*{Acknowledgements}

Many thanks to Lukas Ellinger for his valuable support and guidance during this project.

\clearpage
\bibliography{custom}
\bibliographystyle{acl_natbib}

\clearpage
\onecolumn
\appendix

\section{Link to codebase}
\label{sec:appendix}
The code for our experiments including the datasets is accessible on GitHub at \url{https://github.com/alexandermanev/cross-lingual-llm-consistency}.

\section{Dataset details}\label{sec:app_dataset_details}

\subsection{Multilingual factual benchmark curation}
Our starting point was the \texttt{KLAR} dataset \citep{wang2025lostmultilinguality}. However, manual inspection revealed significant flaws, including incorrect ground-truth objects, translation errors, and missing languages. Consequently, we curated our own dataset by selecting subjects and relations from \texttt{KLAR}, rigorously fact-checking all associated objects against Wikidata (with Wikipedia as a fallback), and discarding the multiple-choice format in favor of open-ended queries to elicit the model's genuine knowledge. We specifically included edge-case subjects, e.g., involving geopolitical disputes or cultural ambiguities where the ``correct'' answer might be contentious. To handle subjects with multiple valid objects, we included separate entries for each.
This resulted in a challenging dataset designed to surface the very inconsistencies we aimed to study.

\subsection{Relation and language selection}
We selected seven relations designed to reveal different facets of LLM knowledge: \textit{country of citizenship, languages spoken, religion, occupation, capital, city of origin,} and \textit{country of origin}. This selection spans standard factual recall to tasks highly susceptible to geographic, gender, or cultural biases. Most importantly, we aimed for relations where multiple correct answers may exist to observe the models' preferences and biases.

To ensure evaluation quality, we restricted the dataset to six languages where we have native or near-native proficiency: \textbf{English}, \textbf{German}, \textbf{Spanish}, \textbf{Bulgarian}, \textbf{Korean}, and \textbf{Swedish}. This selection provides excellent typological diversity, covering Germanic, Latin, Slavic, and Koreanic families.
We used the \textit{DeepL API} for initial translations of the curated relations, but every entry was manually reviewed and aligned with target-language Wikipedia to ensure semantic accuracy---a step that proved crucial for avoiding translation artifacts.
Ultimately, this yielded a dataset consisting of 7 relations across 6 languages (see \S\ref{sec:app_data_format} for data format examples).

\subsection{Dataset format examples}\label{sec:app_data_format}
Examples for data points in English, German, and Spanish:

\subsubsection{Relation: \textit{capital}, language: \textit{English}}
\begin{small}
\begin{verbatim}
{
    "subject": "Azerbaijan",
    "object": "Baku",
    "index": 1000
}
\end{verbatim}
\end{small}

\subsubsection{Relation: \textit{occupation}, language: \textit{German}}
\begin{small}
\begin{verbatim}
{
    "subject": "Halle Berry",
    "object": "Schauspielerin",
    "index": 3000,
    "subject_en": "Halle Berry",
    "object_en": "actress"
}
\end{verbatim}
\end{small}

\subsubsection{Relation: \textit{country of citizenship}, language: \textit{Spanish}}
\begin{small}
\begin{verbatim}
{
    "subject": "Borís Spaski",
    "object": "Francia",
    "index": 5000,
    "subject_en": "Boris Spassky",
    "object_en": "France"
}
\end{verbatim}
\end{small}

\subsection{Contrast-pair format}\label{sec:app_contrast_pairs}
A representative contrast pair from the steering pipeline is shown below.

\begin{small}
\begin{Verbatim}[breaklines=true,breakanywhere=true]
{
    "question": "Which country is Albert Einstein a citizen of?\n\n
                 Choices:\nA. United States\nB. Germany",
    "positive": "B",
    "negative": "A"
}
\end{Verbatim}
\end{small}

\subsection{Generalization dataset construction} \label{sec:app_generalization_construction}
Aligning a model's answer distribution for specific facts does not guarantee a genuine conceptual shift, i.e., adopting the ``worldview'' associated with the given language.
To evaluate whether targeted factual interventions transfer to broader cultural dispositions, we engineered a novel generalization dataset from scratch.

The dataset consists of 50 culturally specific multiple-choice scenarios per target language---German, Spanish, and Bulgarian. To construct a high-quality, diverse set of questions, we employed an LLM-assisted pipeline using Gemini 3.1 Pro, prompting it to generate highly typical---and occasionally stereotypical---questions spanning distinct cultural categories such as dining habits, social etiquette, and legal discrepancies. Every data point was subsequently manually reviewed and refined to ensure the expected target answer was accurate for the respective culture.
Crucially, all queries are formulated in \textbf{English} and utilize second-person scenario framing (e.g., "You sit down for dinner..."). This design choice queries the model's English-language representations \textit{after} an intervention has been applied, testing whether its internal anchor has shifted to the target demographic.

Each entry provides a scenario with two choices: English- and target-centric. For instance, to probe Spanish cultural norms regarding dinner times, the English-centric choice is \textit{6:30 PM or 7:00 PM}, while the target-centric one is \textit{9:30 PM or 10:00 PM} (see \S\ref{sec:app_generalization_dataset} for examples across all languages).
Success on this dataset indicates that the intervention did not merely patch isolated facts, but successfully bridged the representational gap within the model's semantic hub, forcing it to seamlessly verbalize a localized, non-English worldview while operating entirely in English.

\subsection{Generalization dataset examples} \label{sec:app_generalization_dataset}

\paragraph{German example}
\begin{small}
\begin{Verbatim}[breaklines=true,breakanywhere=true]
{
    "index": 10019,
    "category": "Supermarket & Shopping",
    "question": "You have a dozen empty plastic water bottles at home. What is the standard way to dispose of them?\n\n
    Choices:\n
    A. Throw them into your home's standard recycling bin for curbside pickup.\n
    B. Take them back to a reverse vending machine at the grocery store to get your deposit money back.",
    "english_centric": "A",
    "target_centric": "B"
},
\end{Verbatim}
\end{small}

\paragraph{Spanish example}
\begin{small}
\begin{Verbatim}[breaklines=true,breakanywhere=true]
{
    "index": 20012,
    "category": "Daily Routine & Time",
    "question": "It is 3:00 PM on a Tuesday. What is happening in the commercial district of your town?\n\n
    Choices:\n
    A. Many small businesses and shops have locked their doors and closed for a few hours so people can eat a large meal and rest.\n
    B. All businesses and shops are fully open, as the workday runs continuously from morning until evening.",
    "english_centric": "B",
    "target_centric": "A"
},
\end{Verbatim}
\end{small}

\paragraph{Bulgarian example}
\begin{small}
\begin{Verbatim}[breaklines=true,breakanywhere=true]
{
    "index": 30008,
    "category": "Social Greetings & Gestures",
    "question": "Someone asks you a question, and you want to silently tell them 'Yes'. How do you move your head?\n\n
    Choices:\n
    A. Nod your head up and down.\n
    B. Shake your head from side to side.",
    "english_centric": "A",
    "target_centric": "B"
},
\end{Verbatim}
\end{small}

\section{Baseline details}\label{sec:app_baseline_details}

\subsection{Generation}
To test our hypothesis across different scales and architectures, we selected 4 state-of-the-art LLMs ranging from small dense models to large Mixture-of-Experts: Gemma 3 12B Instruct \citep{gemmateam2025gemma3}, gpt-oss-20b \citep{openai2025gptoss120bgptoss20bmodel}, Qwen 3 30B-A3B \citep{yang2025qwen3}, and Llama 4 Maverick Instruct \citep{meta2025llama4} (accessed via the \textit{Gemini} and \textit{Fireworks AI} APIs for obtaining the baseline results).

A major challenge in open-ended generation is reliably measuring a model's confidence. We explicitly rejected token-level probability calculation (computationally infeasible) and multiple-choice formats, as providing a list of candidates introduces bias and fundamentally alters the task from genuine recall to mere recognition. Instead, we relied on \textbf{repeated sampling}.
For each subject-relation pair, we queried the model 10 times (regardless of how many correct answers) at two non-zero temperatures: $0.8$ and $1.2$. To mitigate prompt sensitivity and ensure the model's responses reflect its internal knowledge, each run randomly selected one of five statement-completion prompt templates (e.g., \textit{"Which city serves as the capital of <subject>? The answer is "}; see Appendix~\ref{sec:app_prompt_templates}). It was guided by a system prompt requesting a single answer to avoid list-based responses (see \S\ref{sec:app_system_prompt}) instead of imposing a limit on the number of generated tokens to avoid interfering with the model's natural output and to better simulate a real-world scenario.
The frequency of an answer across these 10 generations served as a proxy for its probability mass.

\subsection{Evaluation pipeline}\label{sec:evaluation_pipeline}
Generated outputs require structured post-processing because exact string matching cannot capture semantic equivalence (e.g., treating \textit{Christianity} as a broader-term answer for \textit{Catholicism}).
We therefore employed GPT-5.1 as the LLM-based evaluation backbone. Our pipeline operates in two stages:
\begin{enumerate}
    \item \textbf{Extraction:} transforms raw model outputs into a canonical English form of the expected entity type (regardless of whether the answer is factually correct---see Appendix~\ref{sec:app_extraction_prompt} for the full policy). This allows us to convert the repeated generations into an empirical answer distribution by frequency, quantifying the model's confidence in its answers.
    \item \textbf{Judgment:} assesses factual correctness and instruction compliance of the model, combining deterministic alias matching with the LLM judge for edge cases (see Appendix~\ref{sec:app_judgment_prompt} for the full policy). While implemented end-to-end to calculate metrics like precision, recall, F1 score, and mean reciprocal rank (MRR), our primary research question concerns the \textit{distribution of answers across languages}, regardless of whether an answer is correct. The judgment layer therefore goes beyond factuality and can support future work on instruction-following and multilingual output quality.
\end{enumerate}

On top of these empirical distributions, we use several distributional metrics. To quantify cross-lingual inconsistency, we compute the \textbf{Jensen--Shannon distance (JSD)} between the answer distributions of different languages for the same subject.
For the subsequent intervention phase, we also track \textbf{Target Probability Mass (TPM)}, evaluating the statistical significance and consistency of these probability shifts using paired \textbf{Wilcoxon signed-rank tests} and the \textbf{Common Language Effect Size (CLES)}.

\subsection{Baseline findings}\label{sec:app_baseline_findings}
The baseline evaluation confirmed that factual behavior is fundamentally unstable across languages. Across the evaluated models, we observed substantial cross-lingual divergence, measured via pairwise JSD between empirical answer distributions. Mean JSD ranged from 0.1258 (Spanish--Swedish) to 0.2818 (Bulgarian--Korean), demonstrating that the same model does not simply produce translated versions of the same answer distribution across languages. 
Furthermore, the strength of the inconsistency depended strongly on the relation. Averaged across language pairs and temperatures, the highest divergence emerged for \textit{occupation} (0.4051) and \textit{city of origin} (0.3972), followed by \textit{religion} (0.3634). In contrast, \textit{capital} showed the lowest one (0.1253). This pattern is intuitively plausible: relations such as capital are usually highly constrained, whereas occupations, religions, and origin admit broader semantic variation, which exposes stronger cross-lingual instability.
This remained consistent across both sampling temperatures, reflecting a systematic shift in the distribution of answers produced across languages rather than sampling noise.

The baseline findings informed the design of our intervention phase, prompting us to refine our focus to \textbf{German}, \textbf{Spanish}, and \textbf{Bulgarian} to ensure a representative spectrum of inconsistency while preserving meaningful linguistic diversity.
As a closely related Germanic language, German serves as a control to test whether targeted interventions can correct factual divergence even when structural and linguistic distance to English is minimal.
Spanish, representing the Romance branch, introduces a uniquely broad, transatlantic cultural footprint, while Bulgarian provides significant structural and orthographic diversity (Cyrillic script, Slavic branch) alongside a culturally distinct Eastern European perspective---a test case for bridging larger representational gaps without breaking the model.
Conversely, we excluded Swedish due to its phylogenetic redundancy with German. We also omitted Korean; while theoretically interesting, its agglutinative morphology and non-Latin script lead to highly disparate tokenization boundaries, which risk confounding the precise alignment of token representations in the residual stream required by CAA.

\section{Prompts}

\subsection{System prompt}\label{sec:app_system_prompt}
For factual generation, the system instruction was language-specific. The English variant was:

\begin{small}
\begin{Verbatim}[breaklines=true,breakanywhere=true]
Respond with exactly one correct answer. If multiple answers are correct, select only one. Do not provide any explanations, reasoning, or additional context.\n
\end{Verbatim}
\end{small}
Equivalent translated variants were used for Bulgarian, German, Spanish, Korean, and Swedish, as defined in the generation pipeline configuration.

\subsection{Factual prompt templates}
\label{sec:app_prompt_templates}

Example for the 5 prompt templates in English for relation \textit{languages}:
\begin{small}
\begin{Verbatim}[breaklines=true,breakanywhere=true]
"What language does <subject> speak? The answer is: <mask>",
"In which language is <subject> fluent? The answer is: <mask>",
"Which language does <subject> use for communication? The answer is: <mask>",
"Which language is spoken by <subject>? The answer is: <mask>",
"What language can <subject> use to communicate? The answer is: <mask>"
\end{Verbatim}
\end{small}

\subsection{Persona prompts}\label{sec:app_persona_prompts}
Prompting used a fixed persona-augmentation template, prepended to the standard system instruction:

\begin{small}
\begin{Verbatim}[breaklines=true,breakanywhere=true]
Adopt the perspective of a typical person born, raised, and currently living in {country}. Your worldview, background knowledge, and behavioral expectations are entirely grounded in {demonym} culture, education, and society. When asked any question—even in English—rely on what a typical {demonym} would consider natural, correct, and factual.
\end{Verbatim}
\end{small}
For the three target languages, the placeholders were instantiated as follows: \textit{(Bulgaria, Bulgarian)} for \texttt{bg}, \textit{(Germany, German)} for \texttt{de}, and \textit{(Spain, Spanish)} for \texttt{es}.

In the factual benchmark, this persona addition was prepended to the standard language-specific factual system instruction. In the generalization benchmark, it was prepended to the instruction requiring the model to answer with exactly one letter (\texttt{A} or \texttt{B}).

\section{Evaluation policies}

\subsection{Extraction policy}\label{sec:app_extraction_prompt}
The extraction stage converts each raw model output into a structured answer representation. It receives the original \texttt{prompt}, the raw \texttt{model\_output}, and the queried \texttt{relation}, and returns two fields:
\begin{itemize}
  \item \texttt{extracted\_object}: the answer phrase explicitly produced by the model, or \texttt{null} if no discernible answer is present;
  \item \texttt{normalized\_object}: a canonical English form of the same answer.
\end{itemize}

The extractor is strictly designed to capture the model's semantic preference rather than factual correctness. It therefore does not judge truth, infer missing information, or repair the answer. It removes trivial wrappers, extracts a single answer, keeps the first valid entity when multiple answers are listed. However, if the output contains explicit step-by-step reasoning or self-correction, it extracts the final concluded entity. Canonicalization is conservative: the answer may be translated into English and minimally adjusted to match the expected relation type, but its fundamental meaning must remain unchanged.
The expected entity types for canonicalization are:
\begin{itemize}
  \item \texttt{capital}: city name
  \item \texttt{city\_of\_origin}: city name
  \item \texttt{country\_of\_origin}: country name
  \item \texttt{country\_of\_citizenship}: country name
  \item \texttt{languages}: language name
  \item \texttt{occupation}: job title
  \item \texttt{religion}: religion name
\end{itemize}

\subsection{Judgment policy}\label{sec:app_judgment_prompt}
The judgment stage evaluates each output for factual correctness, instruction adherence, and linguistic quality. It receives the full evaluation context---\texttt{prompt}, \texttt{model\_output}, \texttt{subject}, \texttt{objects}, \texttt{relation}, and \texttt{target\_language}---and returns a structured analysis including \texttt{is\_correct}, \texttt{object\_match}, \texttt{response\_category}, \texttt{violations}, \texttt{linguistic\_issues}, and a concise \texttt{justification}.

Correctness is determined via \textbf{truth compatibility (entailment)} rather than exact string identity: the LLM judge may use background knowledge to resolve synonyms, translations, and historical equivalents (e.g., mapping ``Kingdom of England'' to ``England''). Specificity is handled via two core rules: 
\begin{itemize}
    \item \textbf{Subset answers (specific $\rightarrow$ general):} if the model gives a more precise answer than the ground truth (e.g., outputting ``Catholicism'' when the truth is ``Christianity''), it is accepted \textit{only} if the judge can verify the specific claim is factually true for the given subject.
    \item \textbf{Superset answers (general $\rightarrow$ specific):} if the model gives a broader answer (e.g., outputting ``Islam'' when the ground truth is ``Sunni Islam''), it is accepted for semantically broad relations (\texttt{religion}, \texttt{occupation}), but strictly rejected as too vague for highly constrained geographic or linguistic relations (\texttt{capital}, \texttt{city\_of\_origin}, \texttt{country\_of\_origin}, \texttt{country\_of\_citizenship}, \texttt{languages}).
\end{itemize}

To disentangle factual accuracy from instruction-following and language generation capabilities, outputs are strictly classified into the following codebook:

\paragraph{Response categories} (assigned via priority hierarchy)
\begin{itemize}
  \item \texttt{refusal}: the model explicitly declines to answer
  \item \texttt{irrelevant}: gibberish or totally unrelated text
  \item \texttt{incorrect\_fact}: a valid entity was extracted but does not match the ground truth (e.g., hallucination)
  \item \texttt{correct\_lax}: the extracted fact is correct, but the output failed a negative constraint (e.g., contained extra text, wrong language, formatting artifacts)
  \item \texttt{correct\_strict}: the fact is correct, in the target language, and contains zero violations
\end{itemize}

\paragraph{Instruction violations}
\begin{itemize}
  \item \texttt{chatty\_filler}: conversational wrappers or phatic expressions (e.g., ``The answer is...'')
  \item \texttt{excessive\_explanation}: unrequested definitions, dates, or historical context
  \item \texttt{multiple\_answers}: providing a list although a single answer was requested
  \item \texttt{self\_correction}: internal thought traces or challenging the premise of the prompt
  \item \texttt{formatting\_artifact}: leaking internal tokens, JSON structures, or Markdown syntax
  \item \texttt{none}: perfect instruction compliance
\end{itemize}

\paragraph{Linguistic quality}
\begin{itemize}
  \item \texttt{wrong\_language}: the output reverts to English or another non-target language
  \item \texttt{mixed\_script}: words containing characters from different alphabets (e.g., Latin characters in Cyrillic)
  \item \texttt{cross\_lingual\_contamination}: the syntax is correct, but valid words from another language are inserted
  \item \texttt{lexical\_hallucination}: non-existent words, morphological failures, or phonetically plausible gibberish 
  \item \texttt{grammar}: general syntax errors or grammatical failures
  \item \texttt{none}: perfect linguistic usage
\end{itemize}

\section{Intervention implementation details}

\subsection{Model selection rationale}\label{sec:app_model_selection}
To support the rigorous demands of our intervention experiments---repeated sampling, extensive hyperparameter sweeps for activation engineering, iterative adapter training---we required a capable open-weights model within a feasible compute budget.
While our baseline included large frontier models, their parameter counts rendered extensive intervention experiments computationally prohibitive. Pilot screening via our judgment pipeline revealed that \texttt{gpt-oss-20b}, a lighter alternative, suffered from severe multilingual degradation and poor instruction adherence.
Consequently, we selected Gemma 3 12B for all experiments. It provided the optimal trade-off: strong multilingual comprehension, instruction following, and crucially, a stable and measurable baseline of cross-lingual factual inconsistency.

\subsection{CAA}\label{sec:app_caa_computation}

\begin{figure}[h!]
    \centering
    \includegraphics[width=0.48\textwidth]{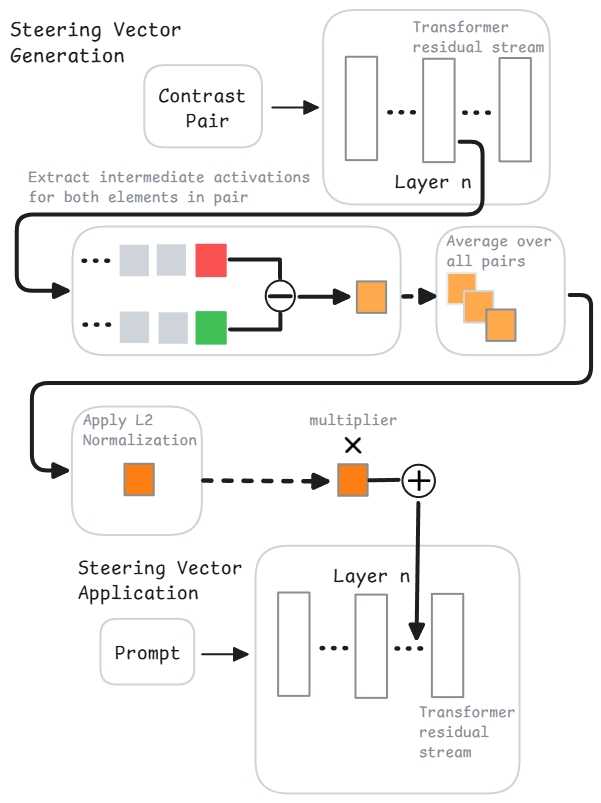}
    \caption{The two-stage process of our CAA pipeline.}
    \label{fig:app_steering}
\end{figure}

\subsubsection{Steering vector computation}
Given the contrast pairs defined in \S\ref{sec:caa}, we computed steering vectors in the standard CAA fashion.
For each pair, we ran the prompt twice through the model: once with the target-language-preferred option as the positive continuation and once with the English-preferred option as the negative continuation. At a chosen intermediate layer $L$, we extracted the hidden state activations corresponding to the answer token and formed a difference vector between the positive and negative activation vectors.
A single, robust steering vector for layer $L$ was then obtained by averaging these difference vectors over all contrast pairs:
\begin{equation}
\mathbf{v}_L = \frac{1}{|\mathcal{D}|}
\sum_{( \mathbf{p},\,\mathbf{c^{+}},\,\mathbf{c^{-}}) \in \mathcal{D}}
\left[ \mathbf{a}_L(\mathbf{p}, \mathbf{c^{+}}) - \mathbf{a}_L(\mathbf{p}, \mathbf{c^{-}}) \right]
\end{equation}
where $\mathbf{a}_L$ is the activation at layer $L$, $c^{+}$ is the target-language-preferred completion, $c^{-}$ is the English-preferred completion, $p$ is the prompt, and $\mathcal{D}$ the contrast-pair dataset.

As a final step, all computed steering vectors could be L2-normalized. This standardization ensures that the magnitude of different vectors is comparable, allowing the intervention strength to be controlled solely and fairly by a scalar multiplier at inference time.

\subsubsection{Steering vector application}

Once computed, the steering vector for a given layer can be used to influence the model's output during inference. We provided the model with an open-ended prompt (e.g., \textit{"What is the country of citizenship of Lionel Messi? The answer is: "}). During the forward pass, we added the steering vector to the model's residual stream activations at the chosen layer. As in the original CAA paper, we applied this addition at every token position generated \textit{after} the prompt. The steering vector's influence is scaled by a multiplier, a key hyperparameter discussed in \S\ref{sec:app_steering_heatmaps}.

\subsubsection{Steering hyperparameter tuning}\label{sec:app_steering_heatmaps}
\begin{figure}[h!]
  \centering
  \includegraphics[width=\linewidth]{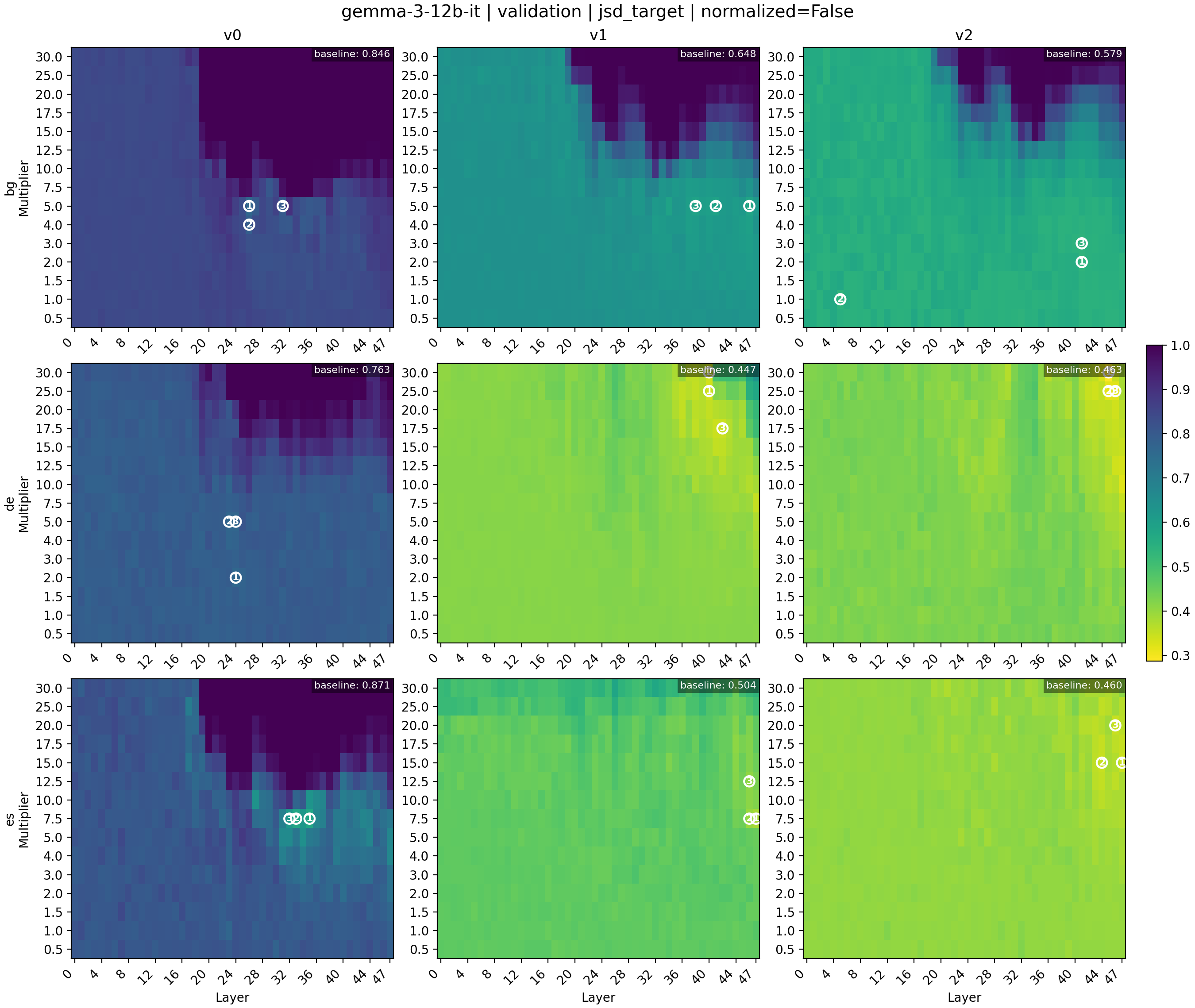}
  \caption{Validation heatmaps for steering across layer and multiplier parameters for each target language and contrast version. Brighter cells indicate better alignment (lower JSD to target). Numbered markers (1--3) highlight the top three configurations for each setting, demonstrating that successful steering requires precise, language-specific hyperparameter selection rather than a uniform architectural rule.}
  \label{fig:app_steering_heatmap}
\end{figure}

Figure~\ref{fig:app_steering_heatmap} illustrates the systematic sweep conducted strictly on the validation split across all 48 layers of Gemma 3 12B and 15 multiplier values (between 0.5 and 30), for all three contrast versions (\texttt{v0}, \texttt{v1}, \texttt{v2}). 
Notably, the strict-disjoint one \texttt{v0} produced substantially fewer usable test subjects (averaging 8 per language) compared to \texttt{v1} and \texttt{v2} (averaging 24.7 per language), because subjects with overlapping English and target candidate sets are discarded.

Prior work suggests that mid-to-late layers are typically the most effective, as this is where models tend to represent the abstract, language-agnostic concepts underpinning factual knowledge \citep{wang2025lostmultilinguality}.
The numbered markers in the heatmaps (indicating the top three optimal configurations for each setting) demonstrate that the steerable locus of cross-lingual preference is not localized uniformly across the model, and that the amount of force required to move the English answer distribution toward the target-language distribution depends on how far apart these preference patterns are in the model's activation space.

Because only a narrow set of regions produces strong gains while the surrounding parameter space yields modest or even destructive effects, steering must be treated not as a fixed intervention, but as a highly sensitive family of interventions where the contrast pair strategy, layer, and multiplier must precisely align with the target-language preference signal.

Table~\ref{tab:app_steering_hyperparameters} details the exact optimal layer and multiplier configurations selected on the validation split and subsequently used for all held-out test evaluations.

\begin{table}[h]
\centering
\small
\begin{tabular}{llcc}
\toprule
Target & Version & Layer & Multiplier \\
\midrule
bg & \texttt{v0} & 26 & 5.0 \\
bg & \texttt{v1} & 46 & 5.0 \\
bg & \texttt{v2} & 41 & 2.0 \\
\midrule
de & \texttt{v0} & 24 & 2.0 \\
de & \texttt{v1} & 40 & 25.0 \\
de & \texttt{v2} & 45 & 30.0 \\
\midrule
es & \texttt{v0} & 35 & 7.5 \\
es & \texttt{v1} & 47 & 7.5 \\
es & \texttt{v2} & 47 & 15.0 \\
\bottomrule
\end{tabular}
\caption{Validation-selected layer and multiplier configurations for CAA steering.}
\label{tab:app_steering_hyperparameters}
\end{table}

\subsection{DPO}\label{sec:app_dpo_training}
\subsubsection{Training hyperparameters}

All DPO experiments were implemented with TRL's \texttt{DPOTrainer} and used the same training recipe for both the benchmark-derived and generalization-derived adapters. We fine-tuned Gemma 3 12B Instruct with parameter-efficient LoRA adapters applied to the standard attention and MLP projection modules (\texttt{q\_proj}, \texttt{k\_proj}, \texttt{v\_proj}, \texttt{o\_proj}, \texttt{gate\_proj}, \texttt{up\_proj}, \texttt{down\_proj}). The LoRA configuration used rank $r=16$, $\alpha=32$, dropout $=0.05$, and \texttt{bias="none"}.

Training used \texttt{beta}=0.01, learning rate $1\times10^{-5}$, 5 epochs, per-device batch size 2, and gradient accumulation over 8 steps. We trained in \texttt{bfloat16}, enabled gradient checkpointing, disabled \texttt{use\_cache}, used fused AdamW (\texttt{adamw\_torch\_fused}) with a cosine scheduler and warmup ratio 0.03, and set the maximum sequence length to 768. Evaluation and checkpointing were both performed every 25 steps, with the best checkpoint loaded at the end of training. All runs used seed 42. Following the PEFT-based DPO setup in TRL, we trained without a separate frozen reference model (\texttt{ref\_model=None}). For Gemma 3 specifically, we additionally patched the training forward pass to inject zero \texttt{token\_type\_ids} when absent, as required by the model implementation.

\subsubsection{Training data}
To maintain a strictly controlled methodological comparison, the adapters were trained separately for each target language (and contrast version---for the benchmark-derived one---\texttt{v0}, \texttt{v1}, \texttt{v2}), with the only variable being the source of the preference data. All training data was formatted into preference triples consisting of a \texttt{prompt}, a \texttt{chosen} continuation, and a \texttt{rejected} continuation:
\begin{itemize}
    \item \textbf{Benchmark-derived DPO:} This adapter directly reused the contrastive signal from the steering pipeline. The target-language-preferred answer served as the \texttt{chosen} continuation and the English-preferred answer as the \texttt{rejected} one. This ensured that the adapter and steering interventions were trained on the exact same underlying preference structure.
    \item \textbf{Generalization-derived DPO:} Preference triples were constructed from the independent generalization dataset. Each item contained an English prompt, with the culturally target-centric option serving as the \texttt{chosen} continuation and the English-centric option as the \texttt{rejected} response.
\end{itemize}
Because the training procedure itself remains unchanged, any performance delta between the two adapter families can be attributed primarily to the data source (factual contrast pairs vs. broader generalization examples).

\section{Detailed robustness and transfer results}
\subsection{Robustness}
Table~\ref{tab:safety_full_dataset} presents the detailed quantitative results of the safety evaluation discussed in \S\ref{sec:robustness}. By measuring performance on the unproblematic held-out dataset, it highlights the core trade-off between targeted alignment and collateral factual degradation.

\begin{table}[h]
\centering
\small
\begin{tabular}{lcccc}
\toprule
Method & Mean $\Delta$TPM$\uparrow$ & Mean $\Delta$JSD$\downarrow$ & Mean $|\Delta$TPM$|\downarrow$ & Mean $|\Delta$JSD$|\downarrow$ \\
\midrule
Prompting & +0.0033 & -0.0030 & 0.0041 & 0.0034 \\
Steering & -0.0200 & +0.0181 & 0.0203 & 0.0181 \\
DPO (Ben.) & -0.0003 & +0.0017 & 0.0019 & 0.0019 \\
DPO (Gen.) & +0.0004 & +0.0009 & 0.0011 & 0.0015 \\
\bottomrule
\end{tabular}
\caption{Safety evaluation on the held-out full dataset relative to the baseline English model. Signed metrics ($\Delta$TPM, $\Delta$JSD) indicate the direction of the distributional shift, while absolute metrics ($|\Delta|$ formats) quantify the total magnitude of perturbation. Lower absolute values indicate safer interventions.}
\label{tab:safety_full_dataset}
\end{table}

\subsection{Transfer}
Table~\ref{tab:generalization_results} provides the complete language-by-language breakdown of the conceptual generalization experiments discussed in \S\ref{sec:generalization}. It illustrates a stark mechanistic dichotomy: while prompting successfully induces a broad, global persona shift, the invasive interventions fail to transfer their benchmark-specific alignment to wider cultural contexts.

\begin{table}[ht]
\centering
\small
\begin{tabular}{llccc}
\toprule
Method & Temp & bg & de & es \\
\midrule
\multirow{3}{*}{Baseline}
& $T=0$   & 0.460 & 0.320 & 0.200 \\
& $T=0.8$ & 0.456 & 0.320 & 0.202 \\
& $T=1.2$ & 0.450 & 0.320 & 0.200 \\
\midrule
\multirow{3}{*}{Prompting}
& $T=0$   & 0.860 & 0.860 & 0.860 \\
& $T=0.8$ & 0.860 & 0.860 & 0.864 \\
& $T=1.2$ & 0.860 & 0.854 & 0.872 \\
\midrule
\multirow{3}{*}{Steering}
& $T=0$   & 0.460 & 0.320 & 0.220 \\
& $T=0.8$ & 0.462 & 0.314 & 0.206 \\
& $T=1.2$ & 0.470 & 0.316 & 0.208 \\
\midrule
\multirow{3}{*}{DPO (Ben.)}
& $T=0$   & 0.440 & 0.320 & 0.220 \\
& $T=0.8$ & 0.450 & 0.316 & 0.208 \\
& $T=1.2$ & 0.450 & 0.320 & 0.202 \\
\bottomrule
\end{tabular}
\caption{Target-centric choice rate on the generalization dataset (50 items per language). Deterministic results isolate the top-1 preference via greedy decoding ($T=0$); sampled results report the expected target-centric probability mass over 10 runs per temperature ($T\in\{0.8,1.2\}$). For steering and DPO, the strongest benchmark-selected configurations are applied.}
\label{tab:generalization_results}
\end{table}

\end{document}